\documentclass{article} 
\usepackage{iclr2017_conference,times}
\usepackage{hyperref}
\usepackage{url}
\usepackage[english]{babel}
\usepackage{graphicx}
\usepackage{mathtools}
\usepackage[]{graphicx}
\usepackage{subfig}
\usepackage{amsfonts}

\usepackage[boxed]{algorithm2e}
\usepackage{algorithmic}
\usepackage{babelbib}

\usepackage{algorithmic}

\newcommand{\longeq}{\scalebox{3}[1]{=}}

\title{An Equivalence of Fully Connected Layer and Convolutional Layer}
\author{Wei Ma, Jun Lu \\
Department of Computer Science\\
EPFL, Lausanne\\
\texttt{wei.ma@epfl.ch, jun.lu.locky@gmail.com}
}

\begin{document}

\maketitle

\begin{abstract}
This article demonstrates that convolutional operation can be converted to matrix multiplication, which has the same calculation way with fully connected layer. The article is helpful for the beginners of the neural network to understand how  fully connected layer  and  the convolutional layer work in the backend. To be concise and to make the article more readable, we only consider the linear case. It can be extended to the non-linear case easily through plugging in a non-linear encapsulation to the values like this $\sigma(x)$ denoted as $x^{\prime}$.
\end{abstract}

\section{Introduction} \label{Introduction}
Many tutorials explain fully connected (FC) layer and convolutional (CONV) layer separately, which just mention that fully connected layer is a special case of convolutional layer \citep{zhou2016dorefa}.  \cite{naghizadeh2009multidimensional} comes up with a method to convert multidimensional convolution operations to $1D$ convolution operations but it is still in the convolutional level. We here illustrate that FC and CONV operations can be computed in the same way by matrix multiplication so that we can convert the CONV layers to FC layers to analyze the properties of CONV layers in the equivalent FC layers, e.g. uncertainty in CONV layers \citep{gal2016uncertainty}, \citep{gal2017concrete}, \citep{blundell2015weight}, or we can apply the methods in FC layers into CONV layers, e.g. network morphism \citep{chen2015net2net}, \citep{wei2016network}. The computation of CONV operations in a matrix multiplication manner is more efficient has but needs much memory storage. 

The convolutional neural network (CNN) consists of the CONV layers. CNN is fashionable and there are various types of the networks that derive from CNN such as the residual network \citep{he2016deep} and the inception network \citep{szegedy2015going}. Our work is non-trivial to understand the convolutional operation well. Formally, convolutional operation is defined by Eq~\eqref{convop} for the continuous $1D$ dimension. Here we use $\odot$ to denote the convolutional operation.
\begin{equation}\label{convop}
[g \odot h](t) = \int_{ - \infty}^{ + \infty} g(\tau) h(t-\tau) d\tau.
\end{equation}
The discrete definition of convolutional operation for $1D$ case is given by Eq~\eqref{discrete1d}. 
\begin{equation}\label{discrete1d}
[g \odot h][n] = \sum_{t =  - \infty}^{ + \infty} g(t) h(n-t), \text{where t and n are integers.}
\end{equation}
But in CNN, we often use the discrete $2D$ convolutional operation as shown in Eq~\eqref{discrete2d}. Section \ref{Common explanation on convolutional (CONV) layer} gives an example about convolutional operation in CNN.
\begin{equation}\label{discrete2d}
[A \odot B] [j_1,j_2] = \sum_{k_1} \sum_{k_2} A(k_1,k_2) B(j_1-k_1,j_2-k_2).
\end{equation}

The following sections are organized as follows. Section \ref{Fully connected (FC) layer} shows the details of matrix multiplication in fully connected layer. Then, Section \ref{Common explanation on convolutional (CONV) layer} introduces the common explanations about convolutional operations. Section \ref{Converting convolutional operation to matrix multiplication} demonstrates how to convert the convolutional operation to matrix multiplication. Section \ref{Experiments} shows the result of a simple experiment on training two equivalent networks, a fully connected network and a convolutional neural network.

\textbf{Notation}: In the rest of the note, scalar variables are denoted as non-bold font lowercases, e.g., $c$ and $s$ are scalar values. Matrix and vectors are denoted  by bold font capitals and lowercases respectively. For example, $\mathbf{W} \in \mathbb{R}^{a \times b}$ means a matrix of the shape $a \times b$ and $\mathbf{x} \in \mathbb{R}^{d\times 1}$ means a column vector with $d$ dimensions.

\section{Fully connected (FC) layer} \label{Fully connected (FC) layer}
\begin{figure}[h]
	\centering
	\includegraphics[width=0.7\textwidth]{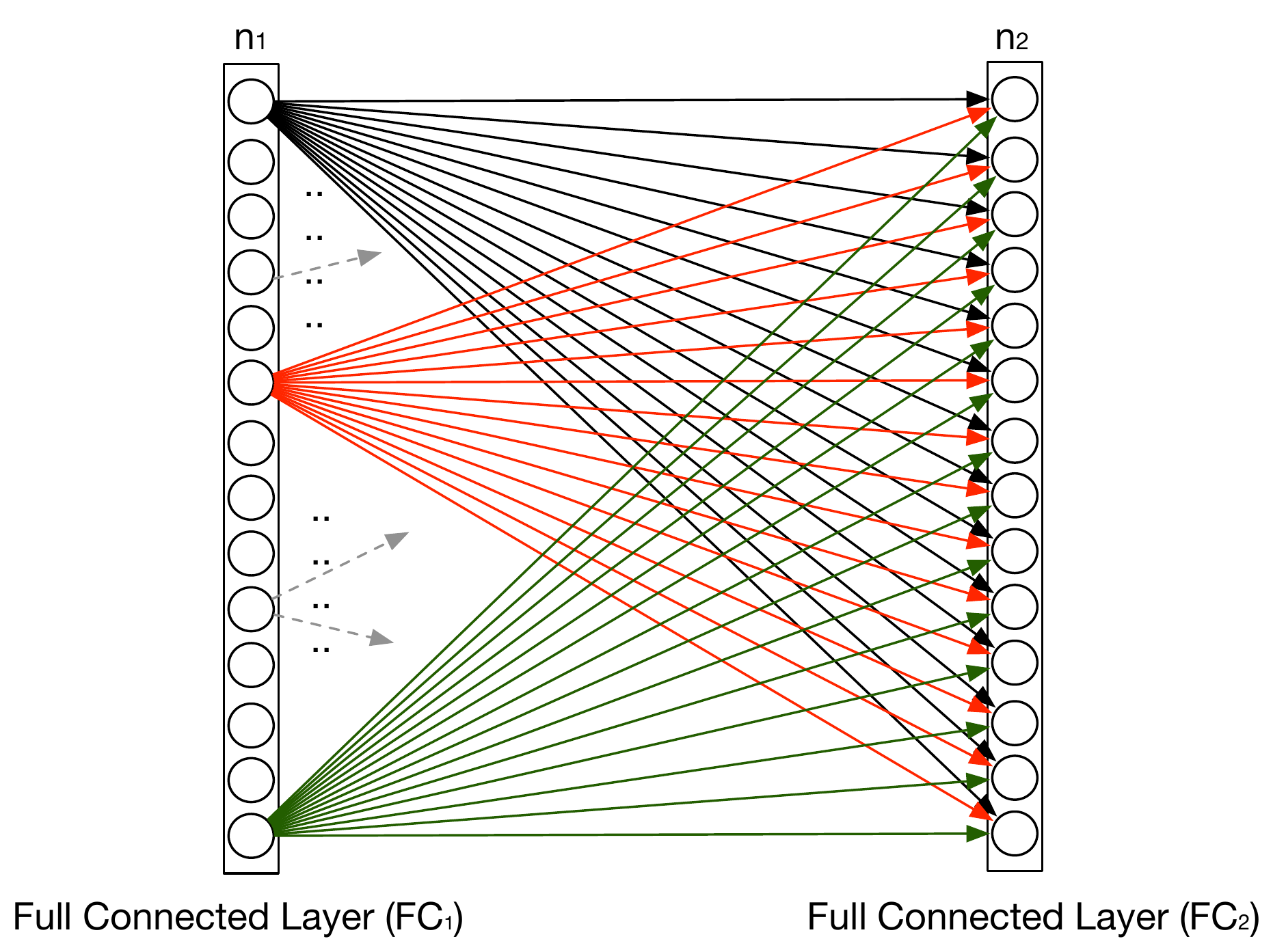}
	\caption{FC layers example with two layers.}
	\label{FullConnectLayer}
\end{figure}

Figure \ref{FullConnectLayer} is a network with two fully connected layers with $n_{1}$ and $n_{2}$ neurons in each layer respectively. The two layers are denoted as $FC_1$ and $FC_2$. Let $\mathbf{x}$ be one output vector of the layer $FC_1$, where  $\mathbf{x} \in \mathbb{R}^{n_1 \times 1}$. Let $\mathbf{W}$ represent the weight matrix of the $FC_2$, where $\mathbf{W} \in \mathbb{R}^{n_1 \times n_2}$ and $\mathbf{w_i}$ is the $i_{th}$ column vector of $\mathbf{W} $. Each column $\mathbf{w_i}$ is the weight vector of the corresponding $i_{th}$ neuron in layer $FC_2$. Thus, the output of $FC_2$ is given by $\mathbf{W}^T \mathbf{x}$.

\section{Common explanation of convolutional (CONV) layer} \label{Common explanation on convolutional (CONV) layer}
\begin{figure}[!ht]
	\centering
	\subfloat[Patch\label{Patch}]{%
		\includegraphics[height=0.3\textwidth]{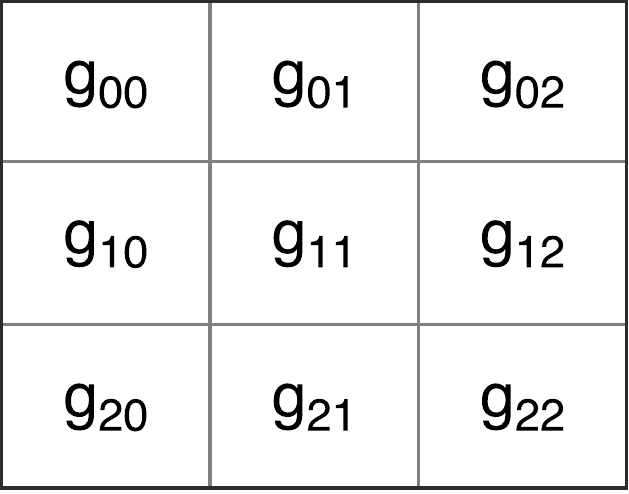}
	}
	\hfil
	\subfloat[Filter\label{Filter}]{%
		\includegraphics[height=0.3\textwidth]{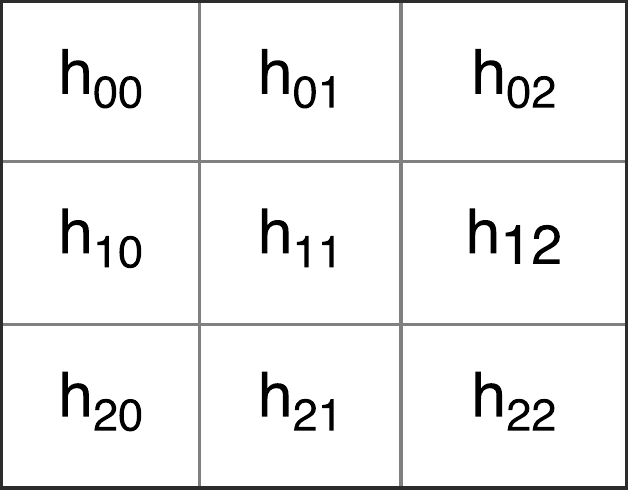}
	}
	\caption{The left image is a patch that is extracted from the input. The right image is a filter. We apply the filter in the whole patch.}
	\label{exampleCONV}
\end{figure}

There are many tutorials on the convolutional operation in deep learning but most of them are unintelligible for the beginners of deep learning. In this section, we illustrate how to understand and compute the convolutional operation in a matrix multiplication manner. Section \ref{Convolution across all the channels} states the common explanations about the convolutional operation. In convolutional operation, point-wise multiplication is often used  for simplicity instead of the convolutional operations shown in Eq~\eqref{discrete2d}. The point-wise matrix multiplication for two variables $i$ and $j$ is shown in the following equation 
\begin{equation} \label{pw}
	(\mathbf{G} \ast \mathbf{H})_{(i,j)}=\mathbf{G}_{(i,j)}\cdot \mathbf{H}_{(i,j)},
\end{equation}  
where $i$ and $j$ are the index, $\mathbf{H}$ is the filter and $\mathbf{G}$ is the input patch (i.e. a patch from the whole input with same shape of the filter). For example, to compute the convolution of $patch$ and $filter$ in Figure \ref{exampleCONV} denoted as $\mathbf{G}$ and $\mathbf{H}$ respectively, according to Eq~\eqref{discrete2d}, 
$\mathbf{G} \odot \mathbf{H}= \sum_{i=0}^{2} \sum_{j=0}^{2} \mathbf{G}_{(i,j)} \cdot \mathbf{H}_{(2-i,2-j)}$. However, in practice, we compute $\mathbf{G} \ast \mathbf{H} = \sum_{i=0}^{2} \sum_{j=0}^{2} \mathbf{G}_{(i,j)} \cdot \mathbf{H}_{(i,j)}$ by point-wise multiplication. The difference between convolution and point-wise multiplication is that convolutional operation needs reverse the filter $\mathbf{H}$ along every dimension.

\subsection{Convolutional operation} \label{Convolution across all the channels}
\begin{figure}[h]
	\centering
	\includegraphics[width=0.8\textwidth]{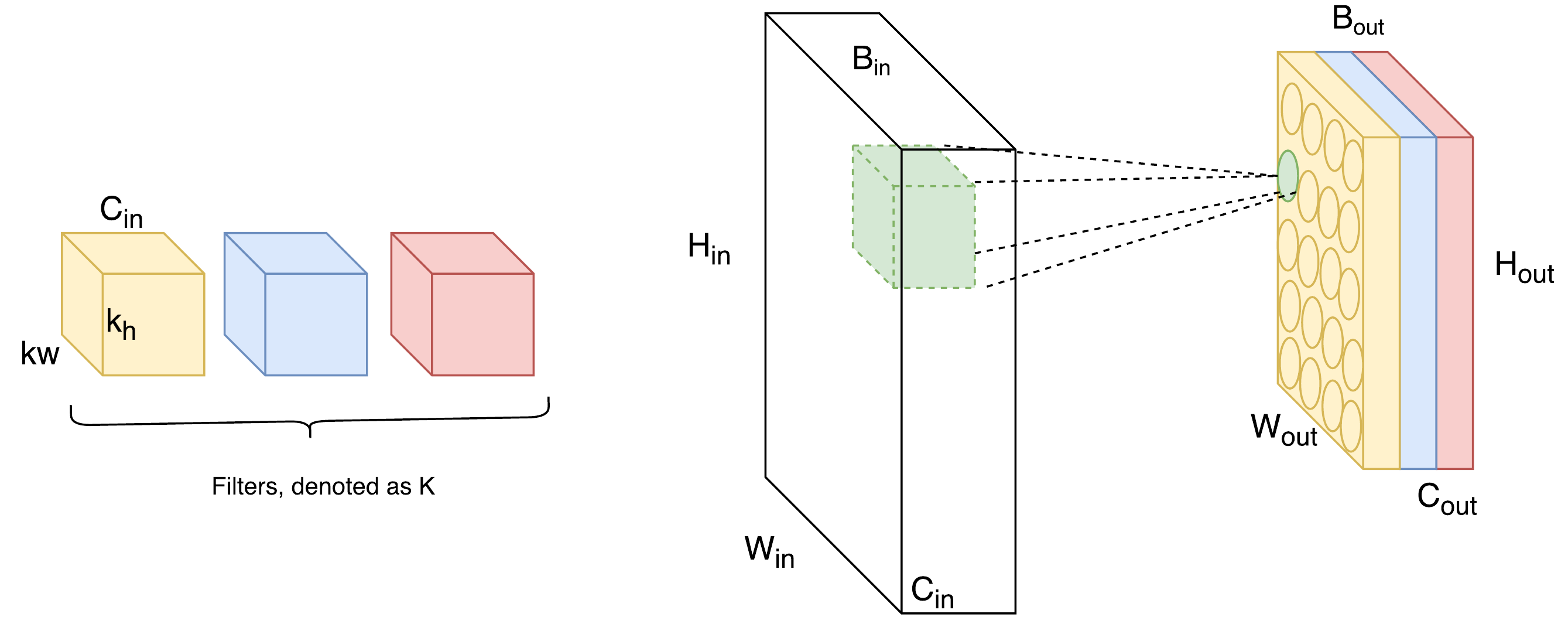}
	\caption{convolutional layer}
	\label{CommonCNN}
\end{figure}

In Figure~\ref{CommonCNN},  $\mathbf{B_{in}}$  and $\mathbf{B_{out}}$ are two $3D$ input and output tensors of $L_1$ CONV layer, where $\mathbf{B_d} \in \mathbb{R}^{H_{d} \times W_{d} \times C_{d}}$, $d \in \{in,out\}$ and $H_{d}$ , $W_{d}$ and $C_{d}$ are height, width and the number of channels respectively. Here we take $3$ filters as an example that is shown in Figure~\ref{CommonCNN}. Every kernel is of size $\mathbb{R}^{k_h \times k_w \times C_{in}}$, where $k_h$, $k_w$ and $C_{in}$ are the height, width and number of channels respectively. We use three different colors (i.e. yellow, blue and red in Figure~\ref{CommonCNN}) to differentiate these three filters respectively. The dashed lines in Figure~\ref{CommonCNN} depict the convolution operation between the yellow filter and the green patch in $\mathbf{B_{in}}$ and its result is put into the corresponding position (green circle) in $\mathbf{B_{out}}$. Every filter moves across $\mathbf{B_{in}}$ from left to right and up to down at a step size (also called as stride number). The different color positions in $\mathbf{B_{out}}$ are the output of the kernel with same color. The process is defined as the convolutional operation in CNN denoted as $\otimes$. Let $\mathbf{K}$ represent the set of the kernels, i.e, $\mathbf{K} \in \mathbb{R}^{f \times k_h \times k_w \times C_{in}}$ where $f$ is the number of the kernels (in our example $f$ is 3). We can denote $\mathbf{B_{out}} = \mathbf{K} \otimes \mathbf{B_{in}}$.\\

\subsection{Relationship between input shape and output shape} \label{Relationship between input shape and output shape}
There exists a relationship between the input shape and output shape in the convolutional operation. Stride  can be denoted as  $s_w$ in the width direction and denoted as $s_h$ in the height direction respectively. Usually, $s_h$ and $s_w$ are set to be the same value so that we can use $s$ to represent $s_h$ and $s_w$. In practice, to get the desired output shape, we often need to pad zeros around the borders of the input. Let $P$ denote the number of rows or columns that we want to pad for each side  (top and bottom, left and right) . There are three main padding ways, non-zeo padding, half-padding and full-padding \citep{dumoulin2016guide}. Eq \eqref{hout}, \eqref{wout} and \eqref{cout} show the relationships between the input shape and output shape of a convolutional operation:
\begin{equation} \label{hout}
	H_{out}= \frac{H_{in}+2P-k_h}{s}+1,
\end{equation}
\begin{equation} \label{wout}
W_{out} = \frac{W_{in}+2P-k_w}{s}+1,
\end{equation}
\begin{equation} \label{cout}
C_{out} = f, 
\end{equation}
where $H_{out}$/$H_{in}$, $W_{out}$/$W_{in}$, $C_{out}/C_{in}$ and $f$ are the output/input height, output/input width, number of output/input channels and the number of filters respectively.
 
\section{Converting convolutional operation to matrix multiplication} \label{Converting convolutional operation to matrix multiplication}

We here extend the analysis of \citet{li2015cs231n} and \citet{gal2016uncertainty} (Section 3.4) and give more details about how to convert a CONV layer into a FC layer. 

\begin{figure}[!ht]
	\centering
	\subfloat[Input Stretch\label{InputStretch}]{%
		\includegraphics[height=0.3\textwidth]{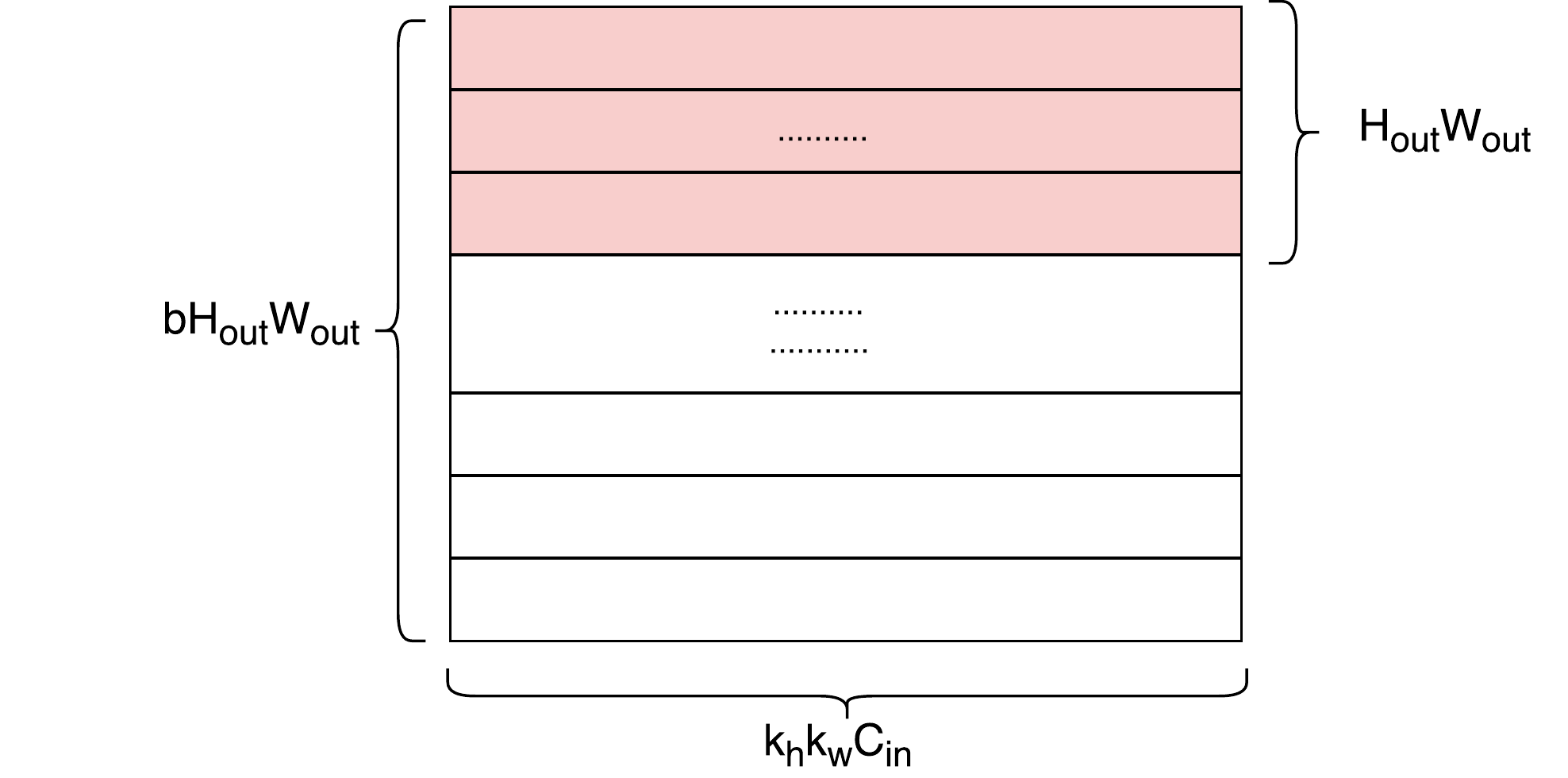}
	}
	\hfil
	\subfloat[Filters Stretch\label{FilterStretch}]{%
		\includegraphics[height=0.3\textwidth]{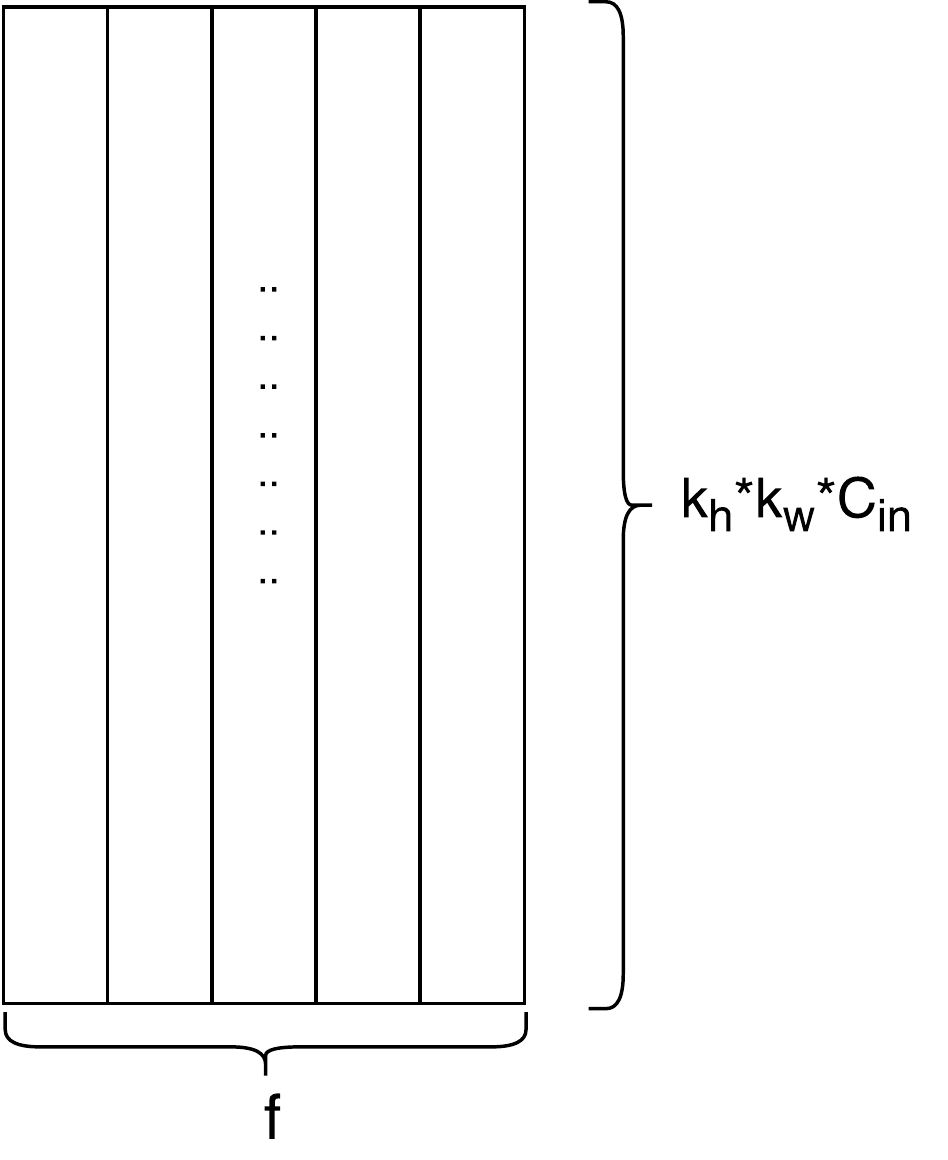}
	}
	\caption{Stretch input and filters}
\end{figure}

We adopt the convolutional view point as shown in Figure \ref{CommonCNN}. We further assume that $H_{in}$ and $W_{in}$ have contained the padding part and the batch size is set to be $b$ (or simply think of $b$ to be the number of samples). The kernel moves across the spatial space $H_{in} \times W_{in}$ by the stride step $s$. It is equal to extracting patches of size $k_h \times k_w \times C_{in}$ according to the movement of the kernel in the input and then the kernel is convolved or point-wise multiplied with the patches. Each patch can be flattened to a row vector with dimension $\mathbb{R}^{1 \times k_h k_w C_{in}}$. These patches constitute a matrix whose dimension is $\mathbb{R}^{(H_{out}W_{out}) \times  (k_h k_w C_{in})}$ that is shown at the red part in Figure \ref{InputStretch}, where $H_{out}$ and $W_{out}$ can be got from Eq \eqref{hout} and \eqref{wout}. This means that each input from the CONV layer can be seen as $H_{out}\cdot W_{out}$ inputs in a FC layer. The whole matrix in the Figure \ref{InputStretch} is denoted as $\mathbf{M}$ with dimension $\mathbb{R}^{(b H_{out}W_{out}) \times (k_h k_w C_{in})}$. 

Accordingly, each filter also can be flattened (stretched) to a column vector of shape $k_h k_w C_{in} \times 1$. Then all the flattened filters make up a filter matrix (i.e. weight matrix in a FC layer) as shown in Figure \ref{FilterStretch}, denoted as $\mathbf{K}$ whose dimension is $(k_h k_w C_{in}, f)$ and $f$ is the number of the filters. The output is given by $\mathbf{MK}$ whose shape is $(b H_{out}W_{out}, f)$. In the end, if we want to convert the output of the matrix multiplication back to the output of a CONV layer, we can reshape the result to be of shape $(b, H_{out}, W_{out}, f)$. \\


\begin{figure}[h]
	\centering
		\includegraphics[height=0.3\textwidth]{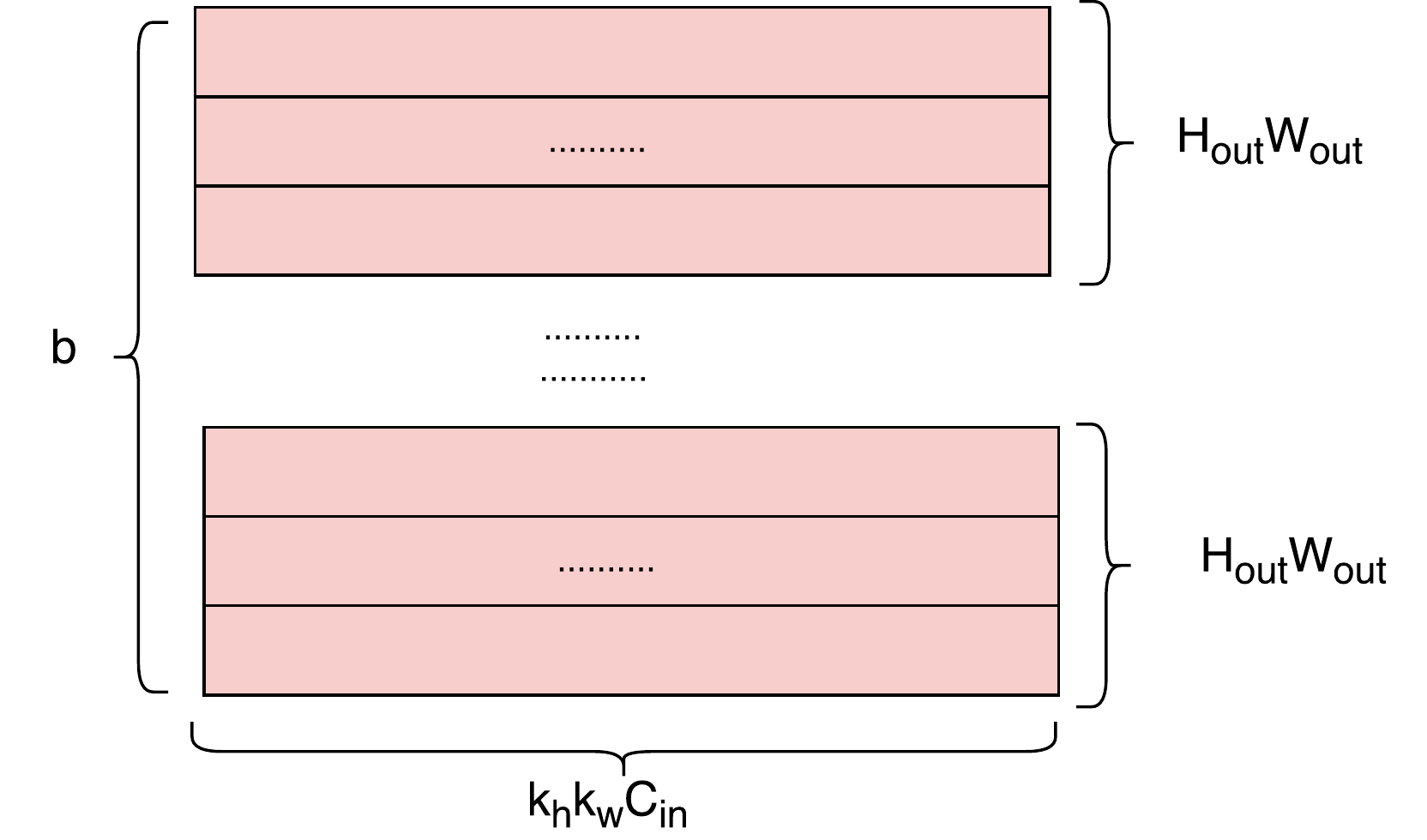}
	\caption{Reshape $\mathbf{M}$ to $3D$ matrix with shape of $(b, H_{out}W_{out}, k_hk_wC_{in})$ denoted as $\mathbf{M^\prime}$}
	\label{3dinputstretch}
\end{figure}

\begin{figure}[!ht]
	\subfloat[Extract pathes from input\label{expathces}]{%
		\includegraphics[width=0.5\textwidth]{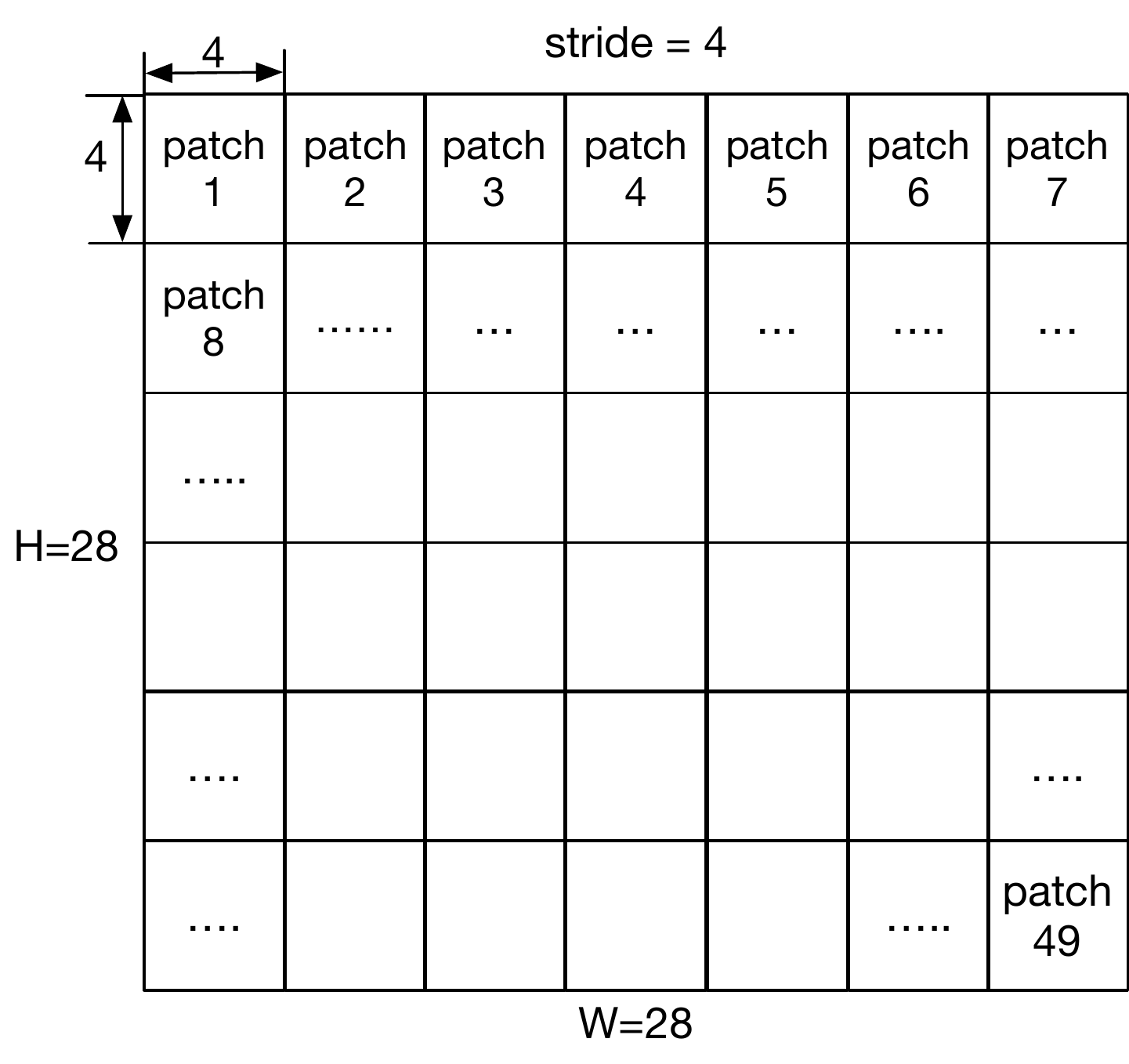}
	}
	\hfill
	\subfloat[Stretching patches\label{exstretch}]{%
		\includegraphics[width=0.4\textwidth]{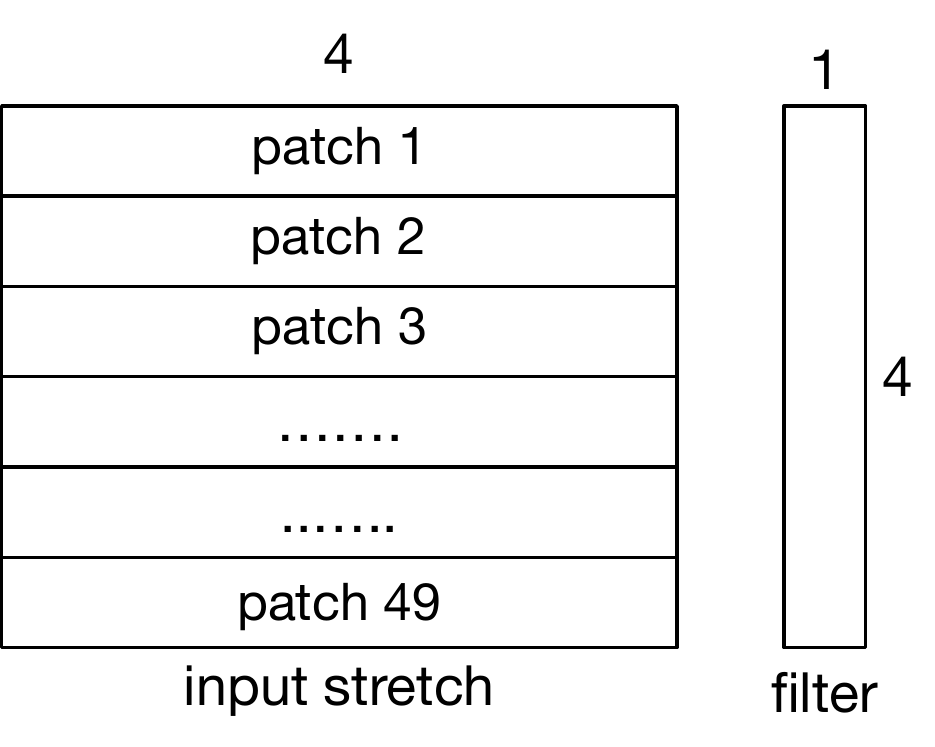}
	}
	\caption{Stretch data example}
	\label{sde}
\end{figure}

For example, we can get an image of shape $(28_{height}, 28_{width}, 1_{channel})$ from MNIST. We use one filter with shape $(4_{height}, 4_{width}, 1_{channel})$ and set the stride to be $4$ to ease the explanation. Figure \ref{sde} demonstrates the stretching process and the result. The patches have no overlap because the width and height of the filter are equal to the stride. We can extract $49$ patches and each patch is of shape $(4, 4)$. Then we flatten each patch to a row vector in Figure \ref{expathces} and stack them vertically together as shown in Figure \ref{exstretch}. The filter is also flattened to a column $4 \times 1$ vector. If there are more than one filter, the flat filters are stacked horizontally.

In our experiment, we use $\mathbf{M^\prime}$ as shown in Figure~\ref{3dinputstretch} instead of $\mathbf{M}$ due to the limitation of APIs of \textit{Keras}. $\mathbf{M}$ is separated to $b$ sub matrices. Each sub matrix is a matrix with shape like the red part in Figure~\ref{InputStretch} with shape $( H_{out}W_{out}, k_h k_w C_{in})$. Then the operation of $\mathbf{MK}$ is divided to the multiplications of sub matrices and $\mathbf{K}$.\\

The process about how to convert convolutional operation to matrix multiplication is described in Algorithm \ref{converConvtoMF}, where we assume $\mathbf{B_{in}}$ has already been padded. We also should notice that the index starts from $0$.

In the deep learning framework, the implementation of converting convolutional operation to matrix multiplication is more efficient by a mapping function of index \citep{vedaldi2015matconvnet}. The method saves memory. The mapping function describes the relationship of the elements in the matrix of stretching patches and in the input matrix. We here don't give a detailed example about this which is out of the scope of the article. To simplify the statement, we assume batch size is equal to $1$, i.e., $b=1$. We know that $\mathbf{M}{(p,q)} \underset{(i,j,d)=t(p,q)} \longeq \mathbf{B_{in}}(i,j,d)$, where $p$ and $q$ are the indexes of $\mathbf{M}$ and $i$, $j$ and $d$ are the indexes of $\mathbf{B_{in}}$. $(i,j,d)=t(p,q)$ is the mapping function of these indexes defined by Eq~\eqref{img2col}, where $ i \in [0, H_{in}-1]$, $i^{\prime} \in [0, k_h-1]$, $i^{\prime \prime} \in [0, H_{out}]$, $ j \in [0, W_{in}-1]$, $j^{\prime} \in [0, k_w-1]$, $j^{\prime \prime} \in [0, W_{out}]$ and $p \in [0, H_{out}W_{out}-1]$ and $q \in [0,k_h k_w C_{in}-1]$. 

\begin{equation} \label{img2col}
	\begin{split}
   \begin{aligned}
	i &= i^{\prime \prime} + i^{\prime}-1, \\ 
	j &= j^{\prime \prime} + j^{\prime} -1, \\
	p &= i^{\prime \prime} + H_{out}(j^{\prime \prime}-1),\\
	q &= i^{\prime} + k_{h}(j^\prime-1) + k_{h}k_{w}(d-1).\\
	\end{aligned}
	\end{split}
\end{equation}

\RestyleAlgo{boxruled}
\LinesNumbered
\IncMargin{2em}
\begin{algorithm}[!ht]
	\KwIn{Feature map $\mathbf{B_{in}}$ with shape (b, $H_{in}$, $W_{in}$, $C_{in}$) after padding\;
		Filters $\mathbf{K}$ with shape ($f$, $k_h$, $k_w$, $C_{in}$)\; 
		Stride $s$\; 
	}
	\KwOut{Feature map, $\mathbf{B_{out}}$ with shape  (b, $H_{out}$, $W_{out}$, $C_{out}$) }
	\BlankLine
	\Begin{
		Create zeros matrix (i.e. the elements in matrix are zero.) $\mathbf{M}$ of shape $(b \cdot  H_{out} \cdot  W_{out}, k_h \cdot  k_w \cdot  C_{in})$\;
		Create zeros matrix $\mathbf{L}$ of shape $(k_h \cdot  k_w \cdot  C_{in}, f)$\;
		Step 1: Compute $H_{out}$ and $W_{out}$ according to Eq~\eqref{hout} and Eq~\eqref{wout}\;
		Step 2: Stretch $\mathbf{B_{in}}$ to $\mathbf{M}$ of shape $(b \cdot H_{out} \cdot  W_{out}, k_h \cdot  k_w \cdot  C_{in})$\;	
		\Begin{
			\For{ j $\in$ range($b \cdot  H_{out} \cdot  W_{out}$) }{
				$l = \lfloor  \frac{j}{(H_{out} \cdot  W_{out})} \rfloor$ \;
				$j \equiv p$ (mod $H_{out} \cdot  W_{out}$)\;
				$m = \lfloor \frac{p}{H_{out}} \rfloor$\;
				$p \equiv t$ (mod $W_{out}$)\;
				$isw = t \cdot  s$\;
				$ish = m \cdot  s$\;
				$\mathbf{M}(j,:) = \mathbf{B_{in}}(l, ish : ish+s, isw:isw+s,:).flatten()$
			}
		}   
		Step 3: Stretch $\mathbf{K}$ to $\mathbf{L}$ of shape $(k_h \cdot  k_w \cdot  C_{in}, f)$\;
		\Begin{
			\For{i $\in$ range($f$)}{
				$\mathbf{L}(:,i) = \mathbf{K}(i, :, :, :).flatten()$
			}
		}   
		Step 4: Compute the output, $\mathbf{B_{out}} =  \mathbf{ML}$\;
		Step 5: Reshape the output to shape of $(b, H_{out},W_{out},f)$\;
		Step 6: Return $\mathbf{B_{out}}$	
	}
	\caption{Converting convolutional operation to matrix multiplication}\label{converConvtoMF}
\end{algorithm}\DecMargin{1em}

\section{Experiments} \label{Experiments}
\begin{figure}[!ht]
	\centering
	\subfloat[Train loss \label{sgdtrainloss}]{%
		\includegraphics[width=0.45\textwidth]{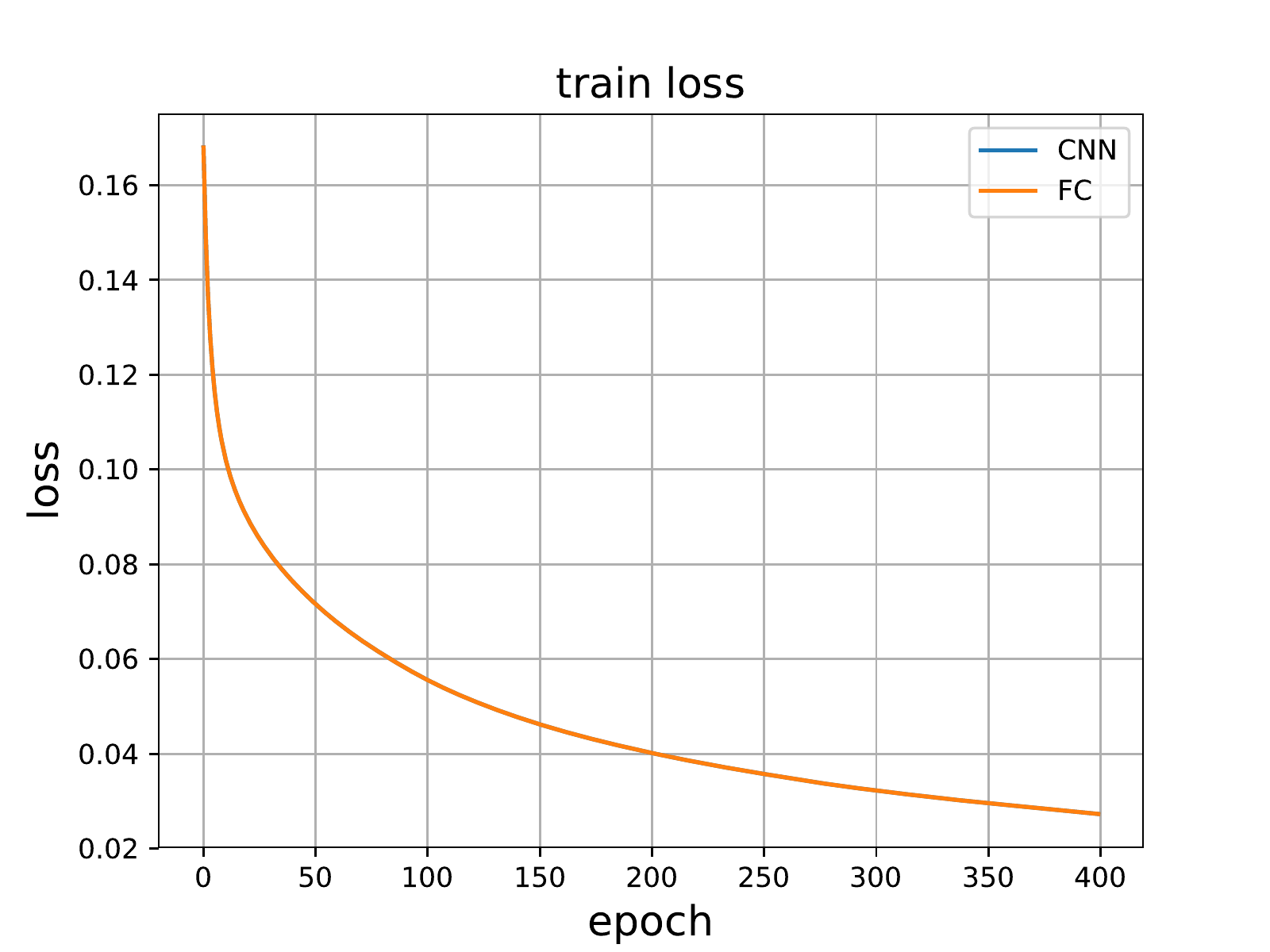}
	}
	\hfill
	\subfloat[Validation loss\label{sgdvalloss}]{%
		\includegraphics[width=0.45\textwidth]{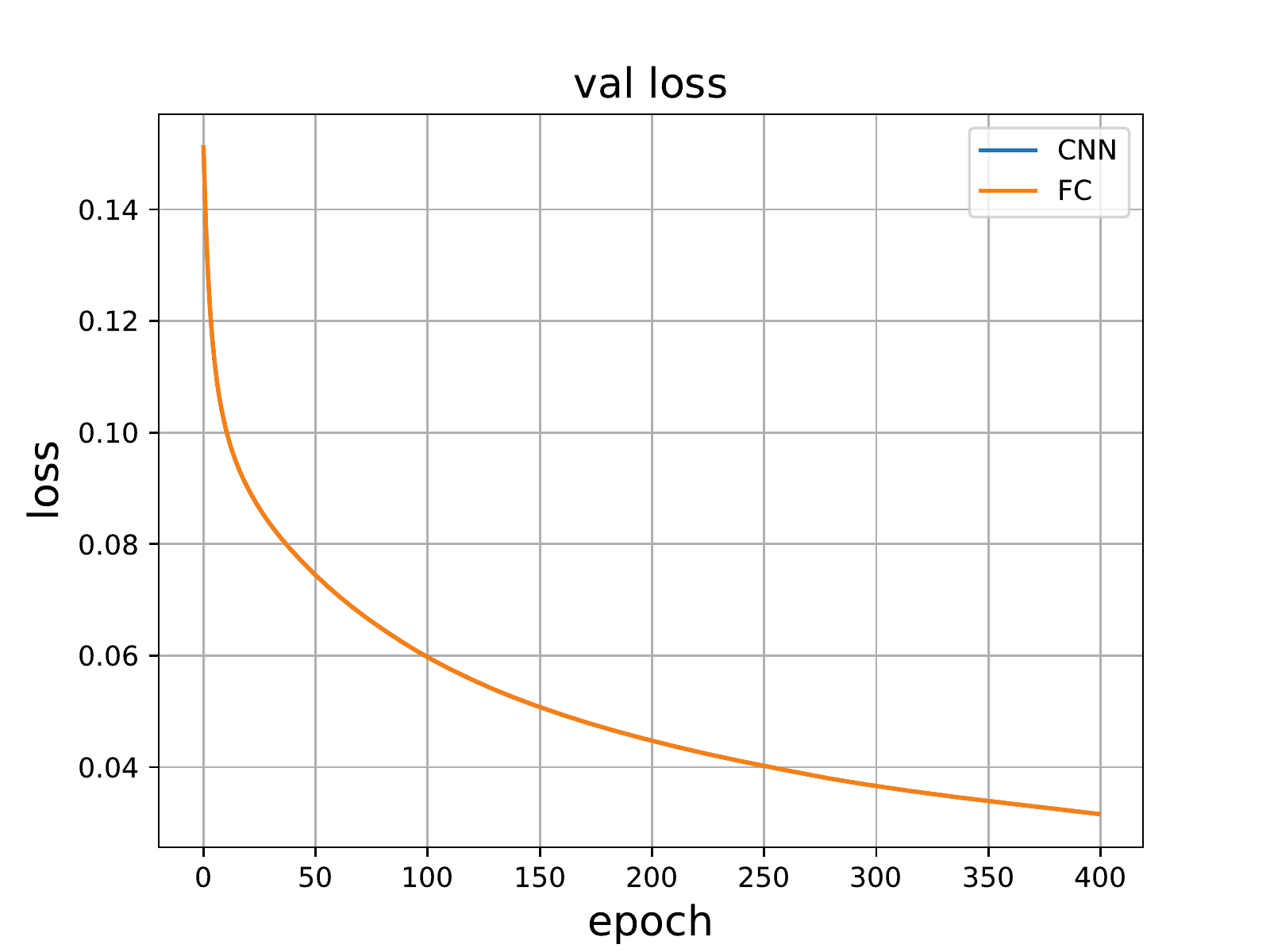}
	}
	\caption{The training and validation loss  curve of the two networks that are optimized by \textit{SGD}}
	\label{sgdloss}
\end{figure}

\begin{figure}[!ht]
	\centering
	\subfloat[CNN \label{}]{%
		\includegraphics[width=0.45\textwidth]{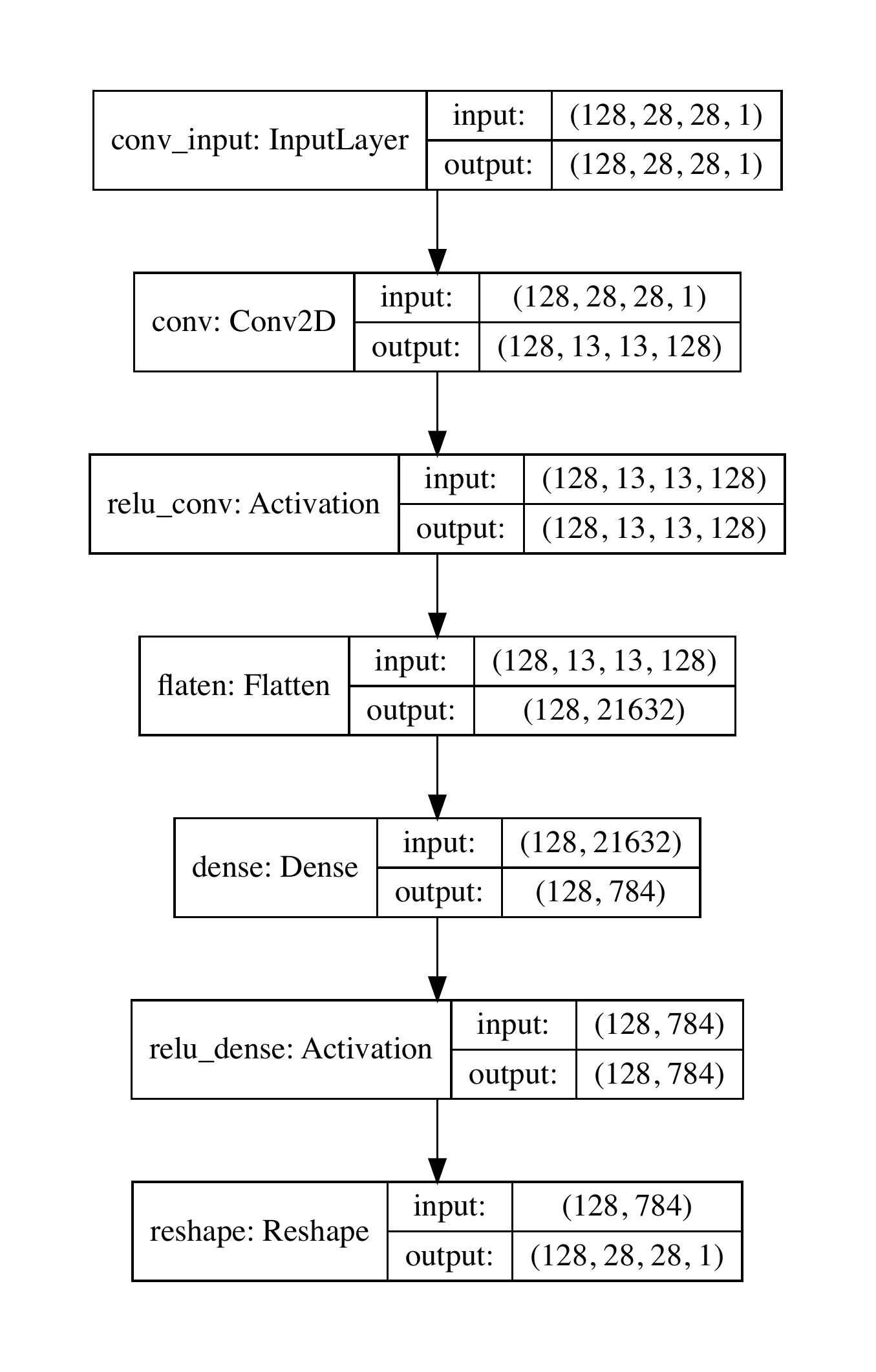}
	}
	\hfill
	\subfloat[FC netowrk\label{}]{%
		\includegraphics[width=0.45\textwidth]{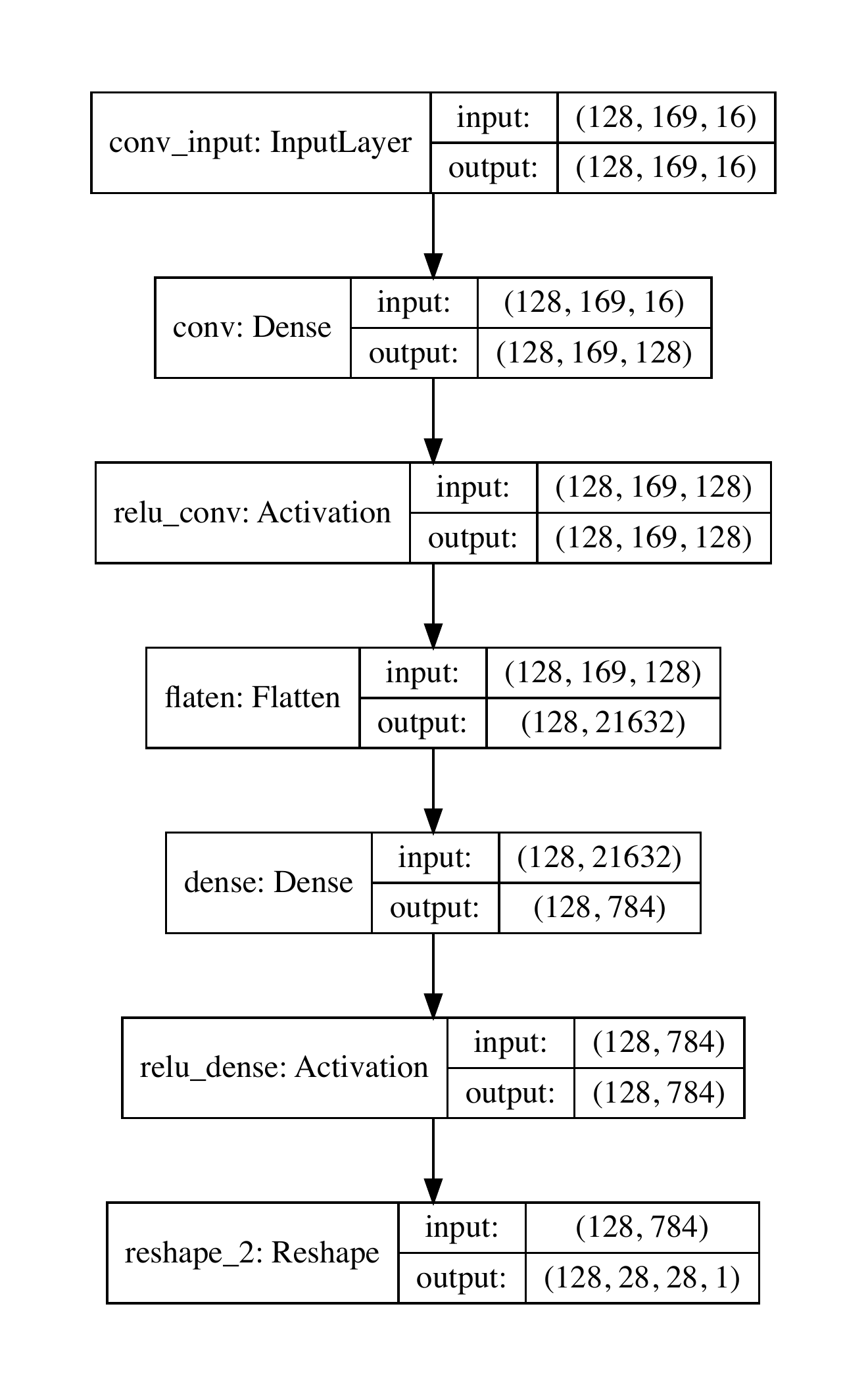}
	}
	\caption{Two networks that are used in the experiments. The first value in the tuple is the batch size. The position of the channels is set to be channel\_last. For more details of the input and output shape, please refer to \textit{Keras} documents.}
	\label{NNofexperimet}
\end{figure}

In the experiment, we use \textit{Keras} to construct one CNN and its equivalent formulation via FC layer (termed as \textit{FC network}) with the same number of the parameters as shown in Figure~\ref{NNofexperimet}. We can ignore flatten, activation layers in Figure~\ref{NNofexperimet}. Both of the networks are going to learn an identity function (i.e. we set the output of the networks to be the original images). For the CONV layer in CNN, the kernel size is set to be $(4,4)$ and the stride is set to be $2$. The difference between the two networks is that the first layer of CNN is a CONV layer whose filter shape is $(4,4,128)$, but the first layer of FC network is a dense layer of whose weight shape is $(16,128)$. We set $batch\_size=128$ and use mean square error (MSE) as the loss function. The optimization method is SGD with $0.01$ learning rate. $1000$ training images and $1000$ validation images are randomly sampled from \textit{MNIST} \citep{lecun2010mnist} as the training data and validation data. The weight initialization method is set acording to \cite{HeZR015}. To train the FC network, the original input data of shape $(1000,28,28,1)$ is converted to the data of shape $(1000, 169, 16)$ based on Algorithm~\ref{converConvtoMF}. We use the same random seed for the two networks so that they have the similar initialization. To simplify the training process, we do not use bias. Both of the two networks are trained for $400$ epochs. The training loss curve and validation loss curve are showed in Figure~\ref{sgdloss}. We can see that the training and validation loss curves of CNN and FC network are almost the same via SGD optimization. We also train the two networks via Adam optimization \citep{KingmaB14}. We compare the results of two optimization. The code $\footnote{Our implementation is available at: \url{https://github.com/statsml/Equiv-FCL-CONVL}}$ is available on Github.\\

One thing we should note as mentioned in Section~\ref{Converting convolutional operation to matrix multiplication}, the input data of $1000$ images for FC network is actually reshaped to $(1000, 169, 16)$ not $(1169000, 16)$. It makes no difference and doesn't affect the weights of the first dense layer in FC network. It just separates matrix $\mathbf{M}$ to $1000$ sub matrices and each sub matrix multiplies the weights $\mathbf{K}$ of shape $(16,128)$ which is the same if we reshape the data to $(1000, 169, 16)$.

\begin{figure}[!ht]
	\centering
	\subfloat[Weight histograms from networks trained by SGD \label{sgdhist}]{%
		\includegraphics[width=0.45\textwidth]{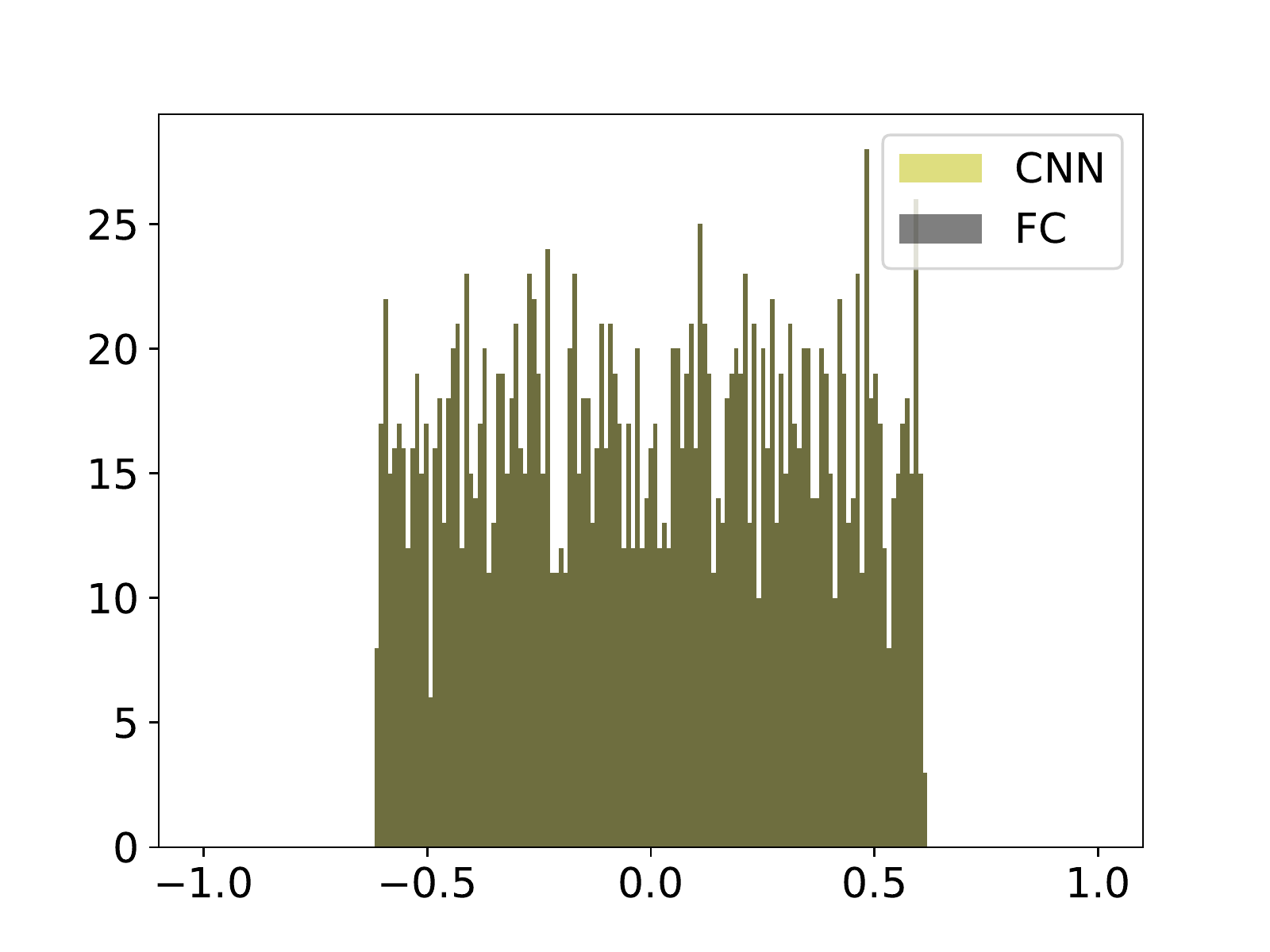}
	}
	\hfill
	\subfloat[Weight histograms from networks trained by Adam \label{adamhist}]{%
		\includegraphics[width=0.45\textwidth]{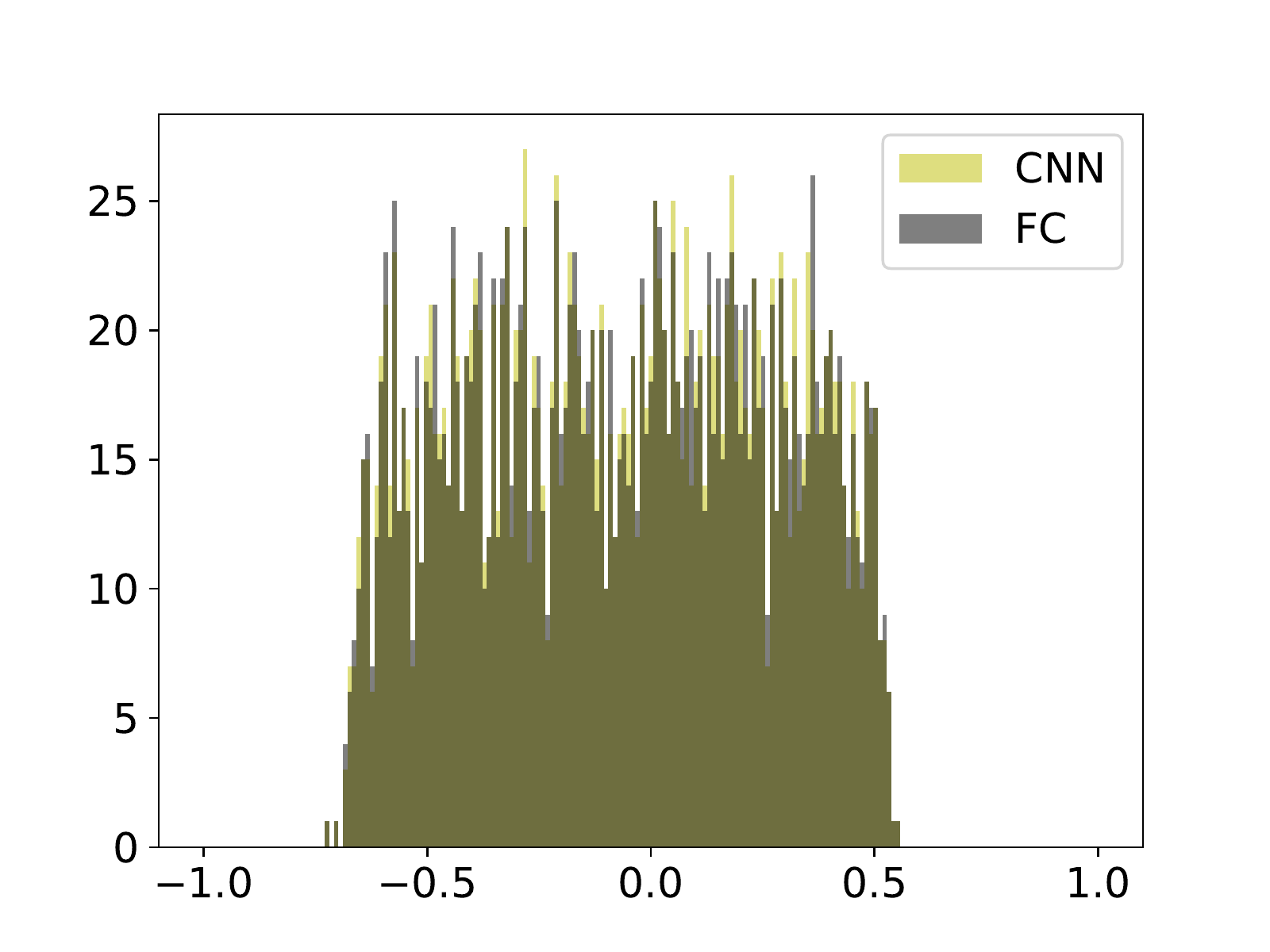}
	}
	\caption{Histograms of filters of the first CONV layer (of shape $(4,4,128)$) and the weights of the first dense layers (of shape $(16,128)$) from CNN and FC network respectively trained by SGD and Adam.}
	\label{histweight}
\end{figure}

\begin{figure}[!ht]
	\centering
	\subfloat[Training loss \label{adamtrainloss}]{%
		\includegraphics[width=0.45\textwidth]{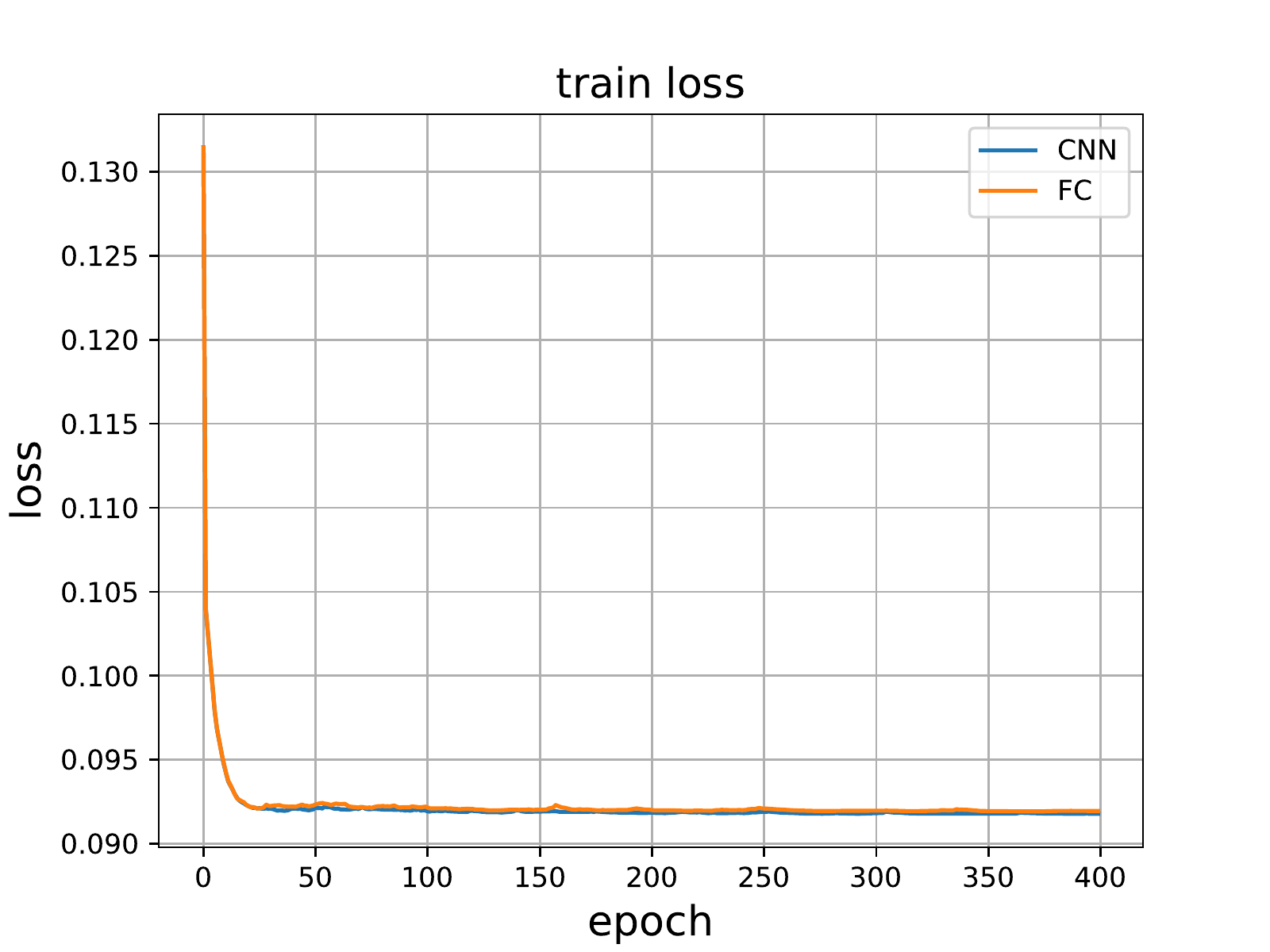}
	}
	\hfill
	\subfloat[Validation loss\label{adamvalloss}]{%
		\includegraphics[width=0.45\textwidth]{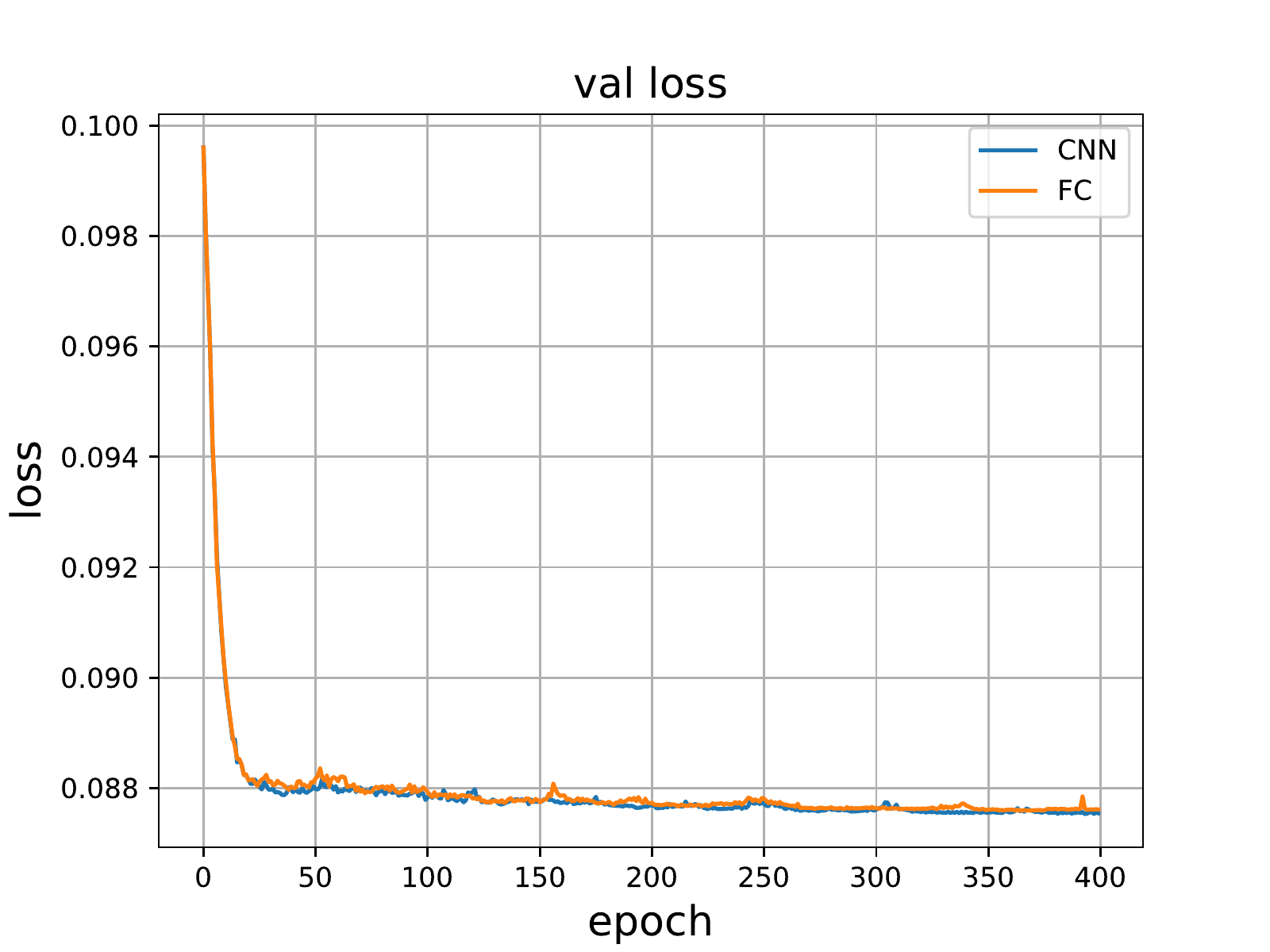}
	}
	\caption{The training and validation loss of the two networks that are optimized by \textit{Adam}}
	\label{adamloss}
\end{figure}

We also extract the outputs of the first layer in CNN and FC network (i.e. the output of \textit{conv} layer in Figure~\ref{NNofexperimet}) denoted by $\mathbf{V}$ and $\mathbf{U}$ for the CNN and FC network respectively. $1000$ new images are used to compute $\mathbf{V}$ and $\mathbf{U}$. Then we compute $\frac{1}{1000}||\mathbf{V} - \mathbf{U}||_F$ and the result is $1.85e-6$. Finally, we plot the histograms of the weights of the two \textit{conv} layers  (denoted as $\mathbf{W_{cnn}}$ and $\mathbf{W_{fc}}$ for CNN and FC network respectively)  as shown in Figure~\ref{histweight}. The histograms from SGD are almost the same for CNN and FC network. And the histograms from Adam almost overlap. We flatten $\mathbf{W_{cnn}}$ and $\mathbf{W_{fc}}$ and their Frobenius norm (F-norm) is $2.12e-7$. We also tried Adam method to optimize the two networks but the training and validation loss curves of the two networks are not overlapped perfectly like Figure~\ref{sgdloss} as shown in Figure~\ref{adamloss}. Adam gets $0.536$ for $\frac{1}{1000}||\mathbf{V} - \mathbf{U}||_F$ and F-norm of its flattened $\mathbf{W_{cnn}}$ and $\mathbf{W_{fc}}$ is $0.0742$. It may be caused by the adaptive learning rates for each parameter that is larger update for infrequent and smaller update for frequent parameters.

\section{Conclusions}
In this note, we illustrate the equivalence of FC layer and CONV layer in the specific condition. Convolutional operation can be safely converted to matrix multiplication, which gives us a novel perspective to understand the convolutional neural network (CNN). And also, in the case where the analysis of CNN is difficult, we can convert the CONV layer in CNN to FC layer and analyze the behavior of CNN in a FC layer manner such as we can analyze the uncertainty in CNN in a FC layer manner.

\pagebreak
\bibliography{iclr2017_conference}
\bibliographystyle{iclr2017_conference}
\end{document}